# Machine Learning in High Volume Media Manufacturing

Siddarth Reddy Karuka[1*], Abhinav Sunderrajan[2*], Zheng Zheng[3#], Yong Woon Tiean[3], Ganesh Nagappan[2*], and Allan Luk[4*]

1 – Seagate Technology, Bloomington, Minnesota, USA
2 – Seagate Technology, Shugart, Singapore
3 – Seagate Technology, Woodlands, Singapore
4 – Seagate Technology, Longmont, Colorado, USA
* These authors are not currently at Seagate
# Corresponding author (zheng.zheng@seagate.com)

**Abstract:** Errors or failures in a high-volume manufacturing environment can have significant impact that can result in both the loss of time and money. Identifying such failures early has been a top priority for manufacturing industries and various rule-based algorithms have been developed over the years. However, catching these failures is time consuming and such algorithms cannot adapt well to changes in designs, and sometimes variations in everyday behavior. More importantly, the number of units to monitor in a high-volume manufacturing environment is too big for manual monitoring or for a simple program. Here we develop a novel program that combines both rule-based decisions and machine learning models that can not only learn and adapt to such day-to-day variations or long-term design changes, but also can be applied at scale to the high number of manufacturing units in use today. Using the current state-of-the-art technologies, we then deploy this program at-scale to handle the needs of ever-increasing demand from the manufacturing environment.

1. Introduction

High vacuum integrity and thin film quality are essential to producing world-class quality and reliable media through multi-layer processing in high throughput sputter process. The sputter process involves a disk moving through a sputter machine that consists of multiple stations, each with their own purpose (Figure 1). A disk carrier carries the disks across the stations such that any given individual disk goes through all the sputter stations sequentially. The stations, depending on the process performed in each individual station, are maintained at a very low pressure or high vacuum. Slot valves separate these chambers from the outside environment and thus help in maintaining the high vacuum. Any degradation or faults in these slot valves lead to small openings that can cause unwanted air to rush into the station chambers, referred to as a leak here after. Such leaks not only lead to wastage of raw material but also affect the quality of the disk, thereby affecting the quality of downline products. These leaks show up in the pressure cycles, where each cycle is defined as the pressure profile observed for each individual station between the time a slot valve is closed and opened again. In other words, a cycle is defined by the time a disk spends undergoing a specific process in a particular station, before moving on to the next station. Catching such minor leak cycles is usually a daunting process and is described in detail in the following sections.

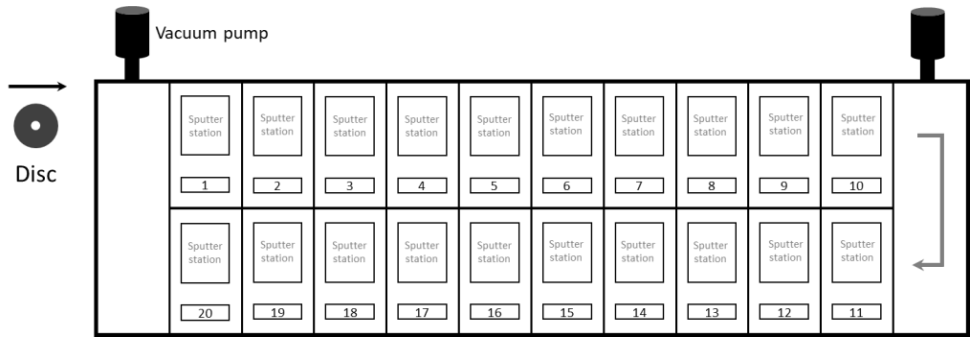

Figure 1: An example sputter machine (details and specifics modified or withheld intentionally)

Since each individual station has their own process or function associated with it, the pressure profile looks different for each of them. However, in general, the stations can be attributed to two types: process and non-process stations. Although the pressure profiles arising from all process and all non-process stations look similar, differences between stations are sufficient to confuse a generic leak detection algorithm. Thus, it is important to have a program that can consider pressure profiles that are characteristic to each individual station. This brings out an important problem the problem of scale. With hundreds of machines on the factory floor, and tens of stations per machine, there are thousands of stations overall. Thus, it is quite difficult to manually develop and deploy an algorithm that can individually scale across each of those stations. This is where our novel unsupervised algorithm, deployed on a scalable Kubernetes cluster, differs from various kinds of algorithms that were previously employed at Seagate. It has also been successfully proven to work on the thousands of stations that are already in use. In addition, the program is also capable of taking in additional units and designs that may be put to production in the future without any changes to the underlying design.

## 2. Materials and Methods

Machine Learning (ML) models can be broadly classified into two categories: supervised and unsupervised models. The main distinction between these categories is the availability of labeled datasets to train the model. A sufficiently large data set with accurate labeling of good and bad data can be used to train a supervised model to learn the characteristics of such data and thus in turn predict any future data to fall into one of those categories. Although we do have labeled data available for the pressure cycles that depict a leak profile, the number of such labeled data is quite low (< 25). Adding on to this issue is the fact that a leak can appear in a fashion that was unknown to us previously and thus any labeled dataset that hasn't seen such a leak profile previously will not be able to accurately catch a new type of leak. The addition of new machines or changes in sputter design processes can also render a previously learned supervised model ineffective and would require the entire labeling and training process to be redone completely. For these reasons, we chose the unsupervised machine learning option, as it not only allows us to develop a model without labeled data but also enables us to adapt to possible variations to the pressure profiles.

The overall program containing both the rule-based decision making and the machine learning portion can be broadly divided into two parts. Initially, a good "baseline" portion of the pressure profile is established immediately after maintenance on a machine. Once a baseline has been established, the machines and stations are continuously monitored for any possible leaks. Figure 2 shows an example of pressure data gathered from a station that just underwent maintenance and has portions of data assigned to baseline and test data. Details on the baseline and test data portions are discussed in detail in further sections.

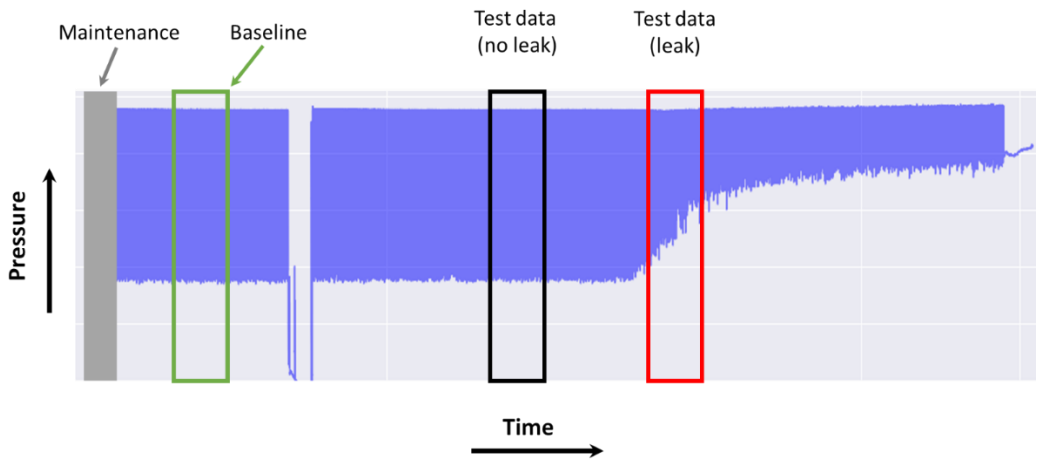

**Figure 2**: Timeline of pressure data from a station, showing maintenance period, baseline and test data portion. The time-series data shown contains thousands of pressure cycles over the period of multiple days, and thus looks like a continuous stream of data. (Certain details in the figure were modified or withheld intentionally)

2.1 Baseline Building

An important assumption that needs to be pointed out here is our reliance on the notion that maintenance results in a non-faulty machine, providing clean pressure profiles indicative of the best performance of that machine. Under this assumption, we consider the pressure profiles observed immediately after to be the "good" ones. To enable such baseline building step, an Apache Airflow scheduler is utilized, that fetches the machine status from Hadoop database every hour and indicates if there was maintenance done recently (Figure 2). A design change would also invalidate the previously built baseline, and thus is also monitored for. An occurrence of either of these triggers a new baseline built for all the stations involved in that machine. In certain cases, the stations take some time, usually less than an hour, to settle down to proper pressure profiles because the machines and stations that were just brought online after maintenance and might be undergoing outgassing or other physical processes. For this reason, we choose to wait for 1 hour before starting to collect baseline data.

Although we believe that the maintenance provides machines and stations that provide clean pressure profiles, it might be possible that some of the pressure cycles are not characteristic of good pressure profiles, mostly caused by sensor instabilities. To account for such pressure cycles and remove them from the baseline data, an outlier removal step is performed. This is when we first introduce machine learning steps in our algorithm, where we use techniques such as Dynamic Time Warping (DTW) [1], Principal Component Analysis (PCA) [2], and Density Based Spatial Clustering of Applications with Noise (DBSCAN) [3].

For the first step of baseline building, data is collected from each station for 1 hour, during which there are hundreds of disks undergoing a sputter process on this station. This data consists of pressure cycles, where each cycle represents one disc undergoing one sputter process in one station. These cycles are separated using the information of slot valves being open and closed, resulting in individual cycles that are then compared to each other using the DTW distance.

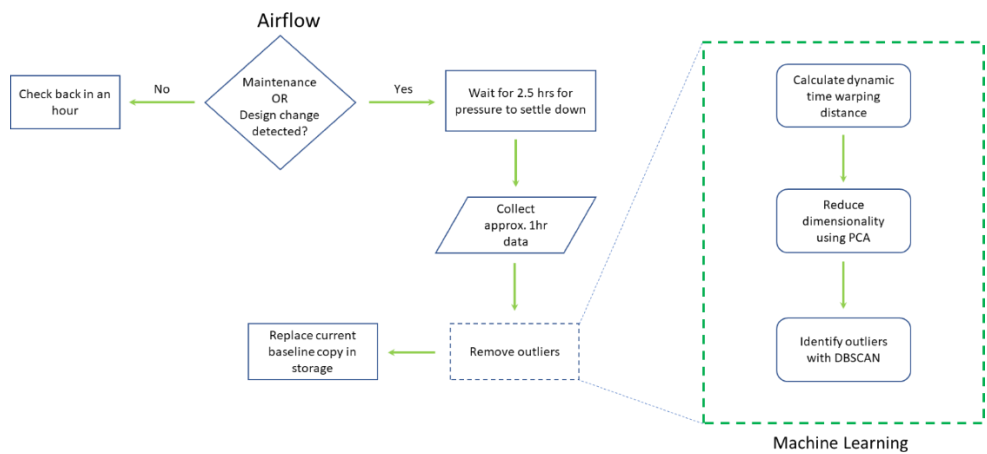

**Figure 3**: Baseline Building

Unlike Euclidean distance or Manhattan distance, which compute distances per point, DTW distance looks at the profile as a whole and thus can consider any lag or lead in the behavior. For this exact reason, DTW is utilized in various scenarios such as speech recognition [4,5], signature matching [6,7], and sign language recognition [8], as they all convey the same meaning despite having slight variations in the actual implementation. For a more relevant example, consider two cycles where the duration is slightly different because of either intentional change in process duration resulting from degraded target sputter source, or unintentional change due to difference in time when the disks are loaded or unloaded. These may cause one of the profiles to appear longer than the other, but the underlying profile still looks the same in shape. In such cases, the DTW distance deduces that these profiles are similar by providing a smaller DTW distance, unlike the Euclidean or Manhattan distance. Since the DTW distances are calculated across all the cycles collected in the hour, the resulting matrix has undesirable high dimensionality, i.e., a multi-hundred-dimensional matrix. PCA is utilized to reduce the dimensionality and capture the most important aspects. PCA projects the given data onto the principal components, where most of the variance can be captured on the projected dimensions as the original dimensions might not be properly aligned to the provided data. Thus, performing PCA on the DTW distance resultant matrix captures 99% of the variance in much fewer dimensions, typically on the order of tens of dimensions. Once low dimensionality is achieved, the data is clustered using DBSCAN, where datapoints are assigned to groups based on how densely they are packed. One of the important parameters for proper functioning of DBSCAN is the estimation of epsilon, a parameter that represents how close the datapoints need to be considered from the same group. Since we perform DBSCAN for different stations independently, it is not possible to fix such a parameter as the behavior changes between stations.

To automate exploration and deciding on an epsilon parameter, we utilize previously explored tools [9,10] that further utilize machine learning techniques such as nearest neighbors and finding the elbow point from the curvature of distance across the data points from the neighbors. The resultant clusters automatically flag outliers, which are then removed from the dataset before storing it to file as the baseline for that station. These baselines are in-place and used for leak analysis while the system keeps checking for further maintenance or design change events every subsequent hour, in which case the stored baseline is deleted, and the entire process is repeated.

2.2   Leak Analysis

Every hour that the algorithm is in use, a status check is performed on all the machines to see if they are "in production". If a machine is in production, a subsequent check is performed to see if there's already a baseline available for the stations of that machine. Once we pass these two stages, a rule-based flowchart (Figure 3), that utilizes machine learning for decision making, allows us to capture a leak in less than an hour. For leak detection, we collect data for 1 hour and perform leak analysis using engineered features and a combination of multiple machine learning algorithms.

The success of any machine learning model depends on the features they use for deciding on the outcome. To make such features as relevant as possible, domain knowledge is a very important tool that can be incorporated in feature engineering. For our algorithm, we perform a two-step leak analysis; first, to check if there is a rise in the "pressure minima", which captures increase in the pressure due to a possible leak, and second, to check if the rise detected in the previous step is due to sensor issues.

For the first step, we chose a set of two features, depicted by "Feature Set 1" (Figure 4). The first of these two features identify a minimum value of each of the pressure cycles in all the cycles collected in one hour of data. Although it is possible for this feature to identify the rise in pressure, the noise in the dataset reduces our confidence in relying solely on this feature alone. To bolster such confidence, we rely on the average pressure observed in the pressure cycles, which happens to occur at different times for process vs non-process stations. Thus, the second feature differs based on the station type, where we consider the mean of the pressure data from the beginning of the cycle to the point where the pressure jumps the maximum for a process station, and the mean of pressure data from the last 20 percent of the data for a non-process station. As shown in Figure 5, both regions correspond to mostly flat regions of the pressure cycle and thus can assist the first feature in establishing a rise in pressure due to a possible leak.

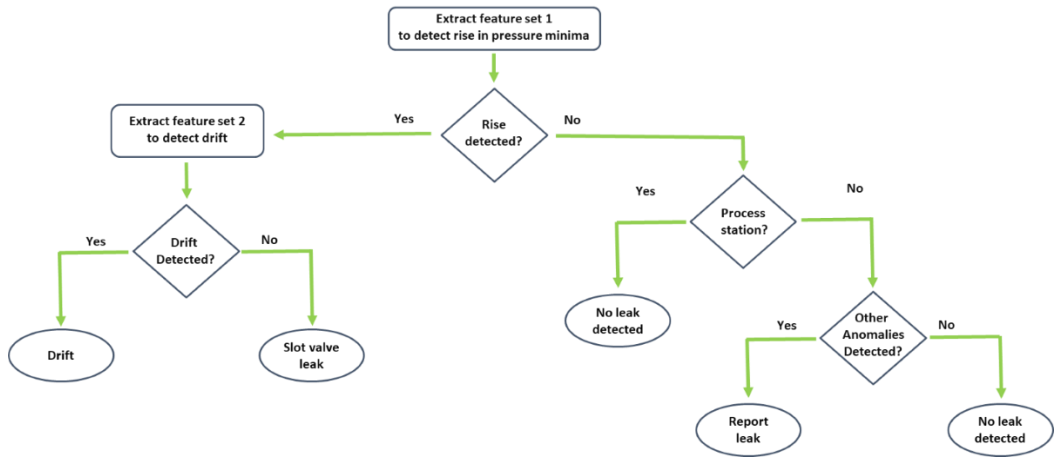

**Figure 4:** Leak Detection Algorithm

Using features from Feature Set 1, a decision needs to be made on whether these features indicate a rise in the pressure due to a possible leak. This is handled by a combination of machine learning tools that we have previously encountered, namely the DTW, PCA and DBSCAN. Similar to what has been done for baseline construction, the outliers from the test dataset are removed before performing leak analysis. We then construct features from Feature Set 1 for both the baseline and the test data set. Density based clustering is then performed on these features to identify the overlap between the baseline and test dataset. To quantify the overlap, we calculate the entropy of the clusters, a parameter that indicates how scattered the data is or how far the datapoints are spread out. Using the entropy for the test dataset, we can further translate it to a leak probability (Figure 6).

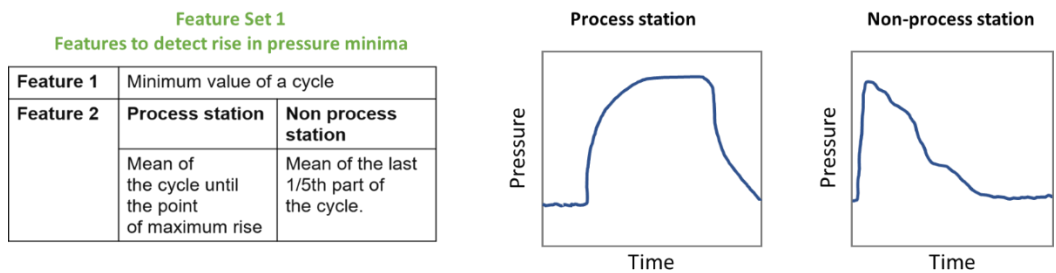

**Figure 5**: Feature Set 1 (details modified or withheld intentionally)

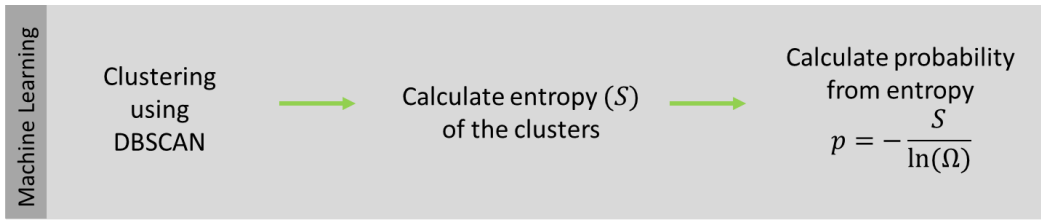

**Figure 6**: Detect Changes

Depending on the leak probability in the first step, the algorithm navigates either of the two possible ways: check for a drift if the leak probability was high, or check for further anomalies, such as rise in pressure maxima in non-process stations, if the leak probability was low.

If the leak probability was high, then we need to check if the signature was from an actual leak or if it was due to sensor behavior. For this purpose, we build another set of features, Feature Set 2, that considers the amplitude of the pressure cycles and how noisy the data is by looking at the difference between successive data points (Figure 7). Using feature set 2, we perform further density-based clustering, similar to what has been done in previous step, to decide on the entropy of the newly formed clusters. From these new clusters and the spread of the test dataset, we can derive a final leak probability (Figure 6), which can decide if the pressure rise observed in step 1 corresponds to an actual leak or to a sensor issue that caused a drift in pressure profiles.

On the other hand, if the leak probability from step 1 is low, we follow the rule-based flowchart from Figure 4 and see if the station is a process or a non-process station. Based on the domain knowledge provided by the sputter team, a process station encounters a rise in pressure minimum for a leak, and thus in the absence of such a rise, we can classify it as a non-leak or good case. However, if the station in consideration is a non-process station, then we perform another check to see if there are any anomalies in the dataset. For this purpose, we utilize the same feature set 2, as changes to the amplitude in non-process stations can also be associated with leaks even when the pressure minima do not change significantly. Similar to what's performed in step 2 after observing a probable leak from step 1, we perform density-based clustering to get to the probability, which then provides us with a final leak probability.

Overall, the final decision is classified into one of the three categories: leak, drift and sensor issue, depending on the leak probabilities obtained over the course of the algorithm. Cases without any issues are captured under the sensor issues category with a very low probability. Based on a threshold set, currently at 85 percent, a notification is sent out to the sputter team if the final leak probability exceeds this set threshold, who can then experimentally verify such a leak.

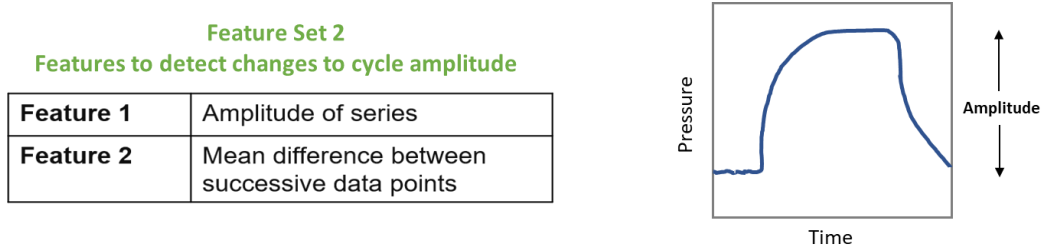

**Figure 7:** Feature Set 2 (details modified or withheld intentionally)

2.3   Deployment

Deployment of the algorithm developed in the previous section carries the same importance, if not more, as the development itself. Since the goal is to monitor thousands of stations in real time, deploying a monolithic application on a regular computer is not advisable for various reasons: a monolithic, or a single large application, has tightly coupled parts in the program and thus a single error can bring the entire application down. Updates to the program require that the whole program needs to be brought down and re-deployed. Any bugs in the program are difficult to catch and address, as the scope of the bug tends to be very broad in a monolithic program. For these reasons, the world is moving towards "microservices" [11], a technique that divides the monolithic program into smaller parts with independent functionality and are usually deployed on a Kubernetes [12] architecture that monitors the programs health and redeploys the services automatically in the case of faults. Any updates can be incrementally rolled out without disturbing the existing services.

We combine various resources available at Seagate to deploy our program at-scale. External tools and services like Apache Airflow, FastAPI, Apache Hadoop, Redis, GitLab, Rancher, OpenSearch are brought together with Seagate internal Kubernetes architecture (Figure 8) for creating a flowchart that executes the program in a smooth and efficient manner (Figure 9).

The Kubernetes architecture includes the training component, HPE Apollo 6500 with 8 Nvidia V100 GPUs, the storage component, Seagate Nytro AFA, and the inference component, HPE Edgeline 4000 with 4 Nvidia P4 GPUs. As the architecture suggests, it is generally used to train and deploy machine learning models that can inference or apply trained models at scale and includes various teams and actions to effectively run a service. This entire architecture is encapsulated under a Kubernetes environment, managed by Rancher, an open-source enterprise Kubernetes management service. For deployment of our program on this platform, we only use the inference component, as our unsupervised anomaly detection model does not allow us to have a trained model, but rather compute the leak probability from ground up at every iteration. Also, for reliability and resilience purposes, these physical servers are also spread out across different physical locations, and thus we have ensured that our program is also ready to deploy on either of these servers at a moment's notice.

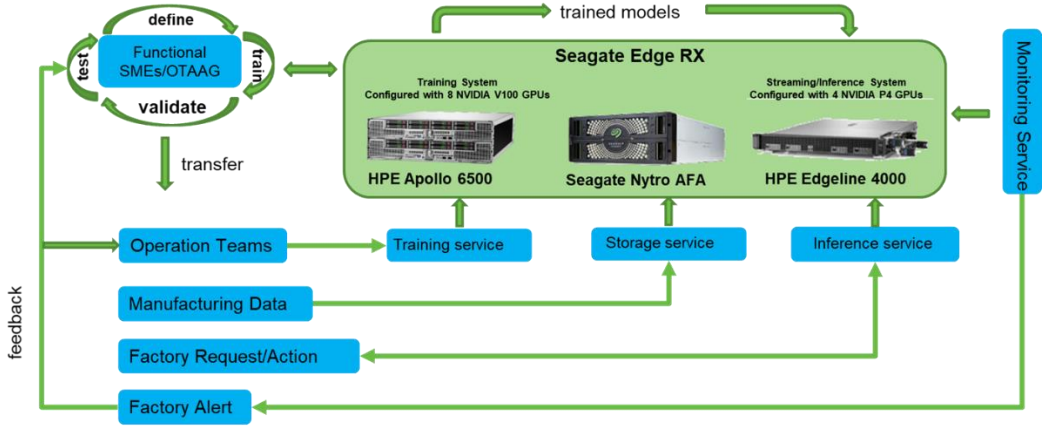

**Figure 8:** Kubernetes architecture

The continuous integration and continuous deployment (CI/CD) of the program to the Rancher environment is enabled by Seagit (Seagate GitLab service), where all our code repositories are versioned and hosted. Any documented changes on GitLab triggers a new CI/CD pipeline that allows the Rancher service to gracefully terminate the old versions without impacting any current running analysis and eventually start up new versions for new incoming analysis.

At the heart of the deployment (Figure 9) is Apache Airflow, a service that schedules various Directed Acyclic Graphs (DAGs) to perform tasks such as monitoring machine status, look for new data, build new baselines, and run leak analysis for each of the hundreds of machines on record. Following is the list of DAGs that are currently active:

1. Fetch data
2. Build Baseline
3. Leak Analysis
4. Email leak report
5. Maintain log database

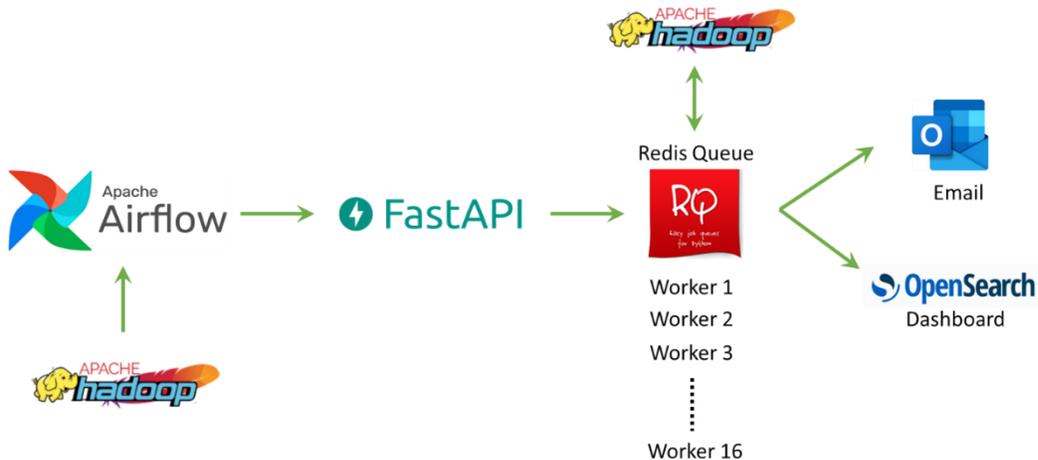

**Figure 9:** Deployment Architecture

One of the first tasks that the DAGs perform is to fetch the machine status, which can take various values such as 'in production', 'maintenance', 'shutdown', etc., from a media data lake managed by Hadoop. This information for each machine is passed onto the Kubernetes cluster, for further tasks, by the FastAPI service. Work on the Kubernetes cluster system is received by Redis, a service that queues jobs and processes them in the background with available workers. Currently we have worker pods, scaling between 8 and 32, which process all the stations simultaneously and perform tasks such as data ingestion, processing, baseline building and leak analysis. Leak analysis results in one of the three outcomes, as discussed previously, which are then stored on the media data lake and OpenSearch dashboard. A different Airflow DAG monitors the data lake for leak results that exceed the threshold and sends out an email notification (Figure 10), the details of which will be discussed in the results section.

An important advantage with the current setup is the ability to incorporate new additions to the production line. Any new machine, independent of the number of stations it has, can be put in through the entire program by adding it to the existing machines file that the Airflow service has. Similarly, increasing or decreasing the amount of workload is automatically handled by Kubernetes, where the number of workers scale between 8 and 32. Thus, the current program not only has unsupervised machine learning to accommodate changes in sputter design, new additions, or daily variations, but the deployment is also engineered to accommodate these fluctuations and the ever-increasing demand in production.

## 3. Results

Deployment of the program, as discussed in the previous section, results in final leak probabilities populated on both the media data lake through Hadoop and on the OpenSearch dashboard. An email notification is sent out if a final leak probability exceeds the set threshold of 85 percent confidence, the format of which is shown in Figure 10.

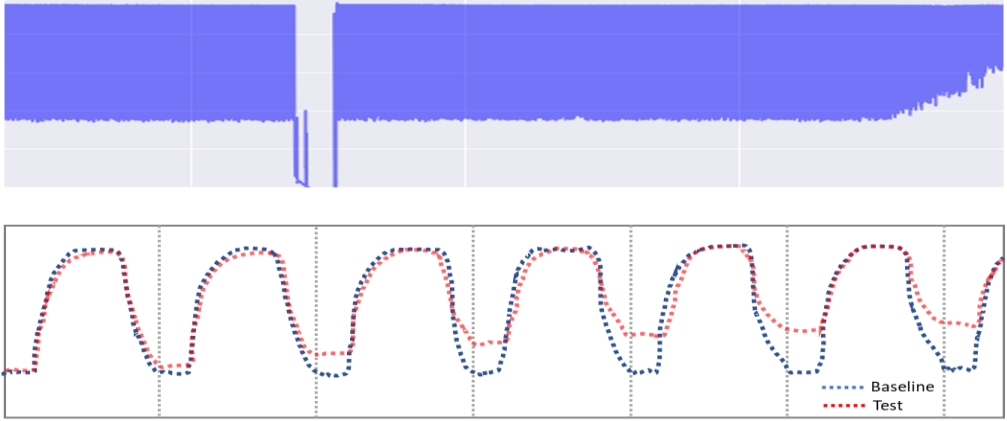

**Figure 10:** Example email notification (details modified or withheld intentionally)

The email notification that is sent out has a multitude of information attached to it, such as the machine and station that's affected, along with information that helps the subject matter expert to diagnose the issue, such as the pressure difference, the probability of a leak, the station type and its function. Finally, we also present figures of the pressure profiles with the email, so that it is easier for the viewer to observe the change in pressure profile. The top one among the two figures shows the pressure profile from the time a baseline has been built to the current time when the leak alert has been sent out, whereas the bottom figure shows a clear distinction between a few cycles selected from the good baseline portion and the current test or analysis portion. In the shown example (Figure 10), the alert was sent out with a 93% probability of a leak, with the top figure showing the leak resulting in loss of vacuum or the increase of pressure in the station, and the bottom figure showing clear difference between the red baseline and blue test data, with blue being higher than red as an indicator of pressure rise or vacuum leak.

On the other hand of the notification system is the OpenSearch toolkit that logs all results of leak analysis. This allows the engineers to view the results without any programming or other technical knowledge to access the Hadoop database. We also built a visual dashboard (Figure 11) that shows the number of probable leaks that were caught in the last 7 days, along with the information on the machine and station. This allows the engineers to identify the most troublesome machine or stations, thus enabling them to assign higher attention to such systems.

Overall, in the time that our program was running in the past few months, we were able to catch 15 of 16 leaks in the second half of FY22. Based on the manufacturing speed of the media, these detections would generally result in avoidance of approximately tens of thousands scrapped or defective media, which leads to savings on the order of millions of dollars per year.

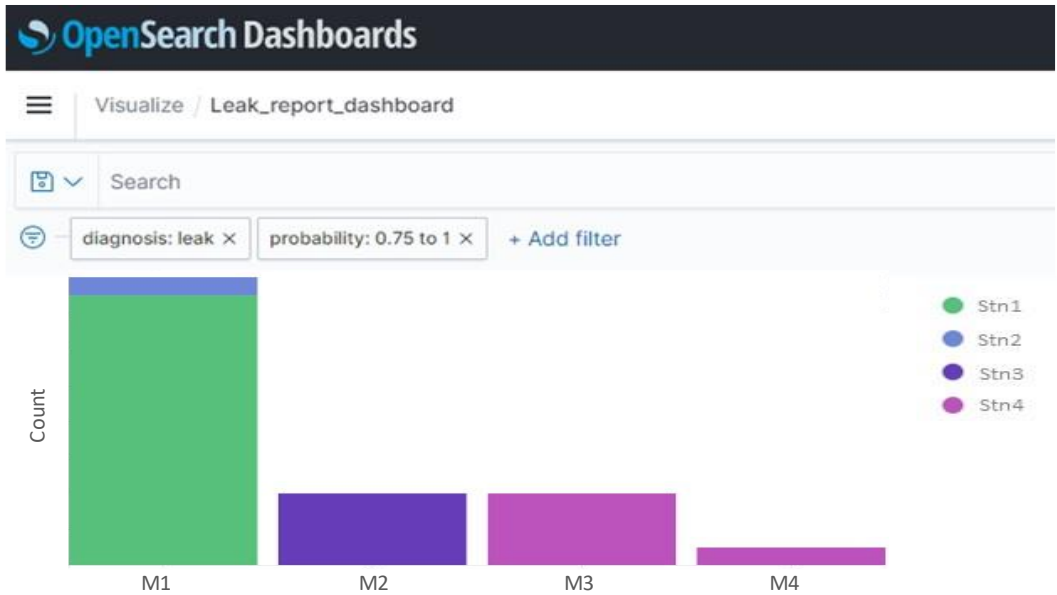

**Figure 11:** Example dashboard display (details modified or withheld intentionally)

## 4. Conclusions

In this paper, we described the development and deployment of an unsupervised machine learning program that detects anomalies or leaks in the sputter vacuum stations. We have detailed the workings of the components involved in the rule-based flowchart that invokes machine learning for decision making. In addition, we have brought together various internal and external components to deploy our algorithm at scale. Combined, these provide us the capability to not only detect anomalies in our current sputter vacuum systems but also keep us ready for any future changes to the manufacturing capabilities. We hope that this article provides an insight into the workings and technological adaptations happening at Seagate's manufacturing environment.

**Author Contributions:** "Conceptualization: Abhinav Sunderrajan (AS), Siddarth Reddy Karuka (SRK), Zheng Zheng (ZZ), Yong Woon Tiean (YWT), Ganesh Nagappan (GN), and Allan Luk (AL); methodology: AS, SRK; software: AS, SRK.; validation: AS, SRK, and ZZ; writing, review and editing: AS, SRK, ZZ.; visualization: SRK.; supervision: GN, AL. YWT. All authors have read and agreed to the published version of the manuscript.

**Acknowledgments:** The authors acknowledge and thank the support from Seagate IT team, Sputter team, and the Data Science Group.

**Conflicts of Interest:** The authors declare no conflict of interest.

**Funding:** No external funding was received for this project. The authors thank Seagate Technology for their support.